# MEDICAL DIAGNOSIS USING NEURAL NETWORK

*S. M. Kamruzzaman, Ahmed Ryadh Hasan[†], Abu Bakar Siddiquee and Md. Ehsanul Hoque Mazumder*

Department of Computer Science and Engineering
International Islamic University Chittagong, Chittagong-4203, Bangladesh
Email: smk_iiuc@yahoo.com, maksud_cse@yahoo.com, sumon_ctg2003@yahoo.com
[†] School of Communication
Independent University Bangladesh, Chittagong, Bangladesh, Email: ryadh78@yahoo.com

**ABSTRACT**

This research is to search for alternatives to the resolution of complex medical diagnosis where human knowledge should be apprehended in a general fashion. Successful application examples show that human diagnostic capabilities are significantly worse than the neural diagnostic system. This paper describes a modified feedforward neural network constructive algorithm (MFNNCA), a new algorithm for medical diagnosis. The new constructive algorithm with backpropagation; offer an approach for the incremental construction of near-minimal neural network architectures for pattern classification. The algorithm starts with minimal number of hidden units in the single hidden layer; additional units are added to the hidden layer one at a time to improve the accuracy of the network and to get an optimal size of a neural network. The MFNNCA was tested on several benchmarking classification problems including the cancer, heart disease and diabetes. Experimental results show that the MFNNCA can produce optimal neural network architecture with good generalization ability.

## 1. INTRODUCTION

Neural networks techniques have recently been applied to many medical diagnosis problems [2] [3] [7] [11]. One of the network structures that have been widely used is the feedforward network, where network connections are allowed only between the nodes in one layer and those in the next layer. Backpropagation algorithm is the most widely used learning algorithm to train multiplayer feedforward network and applied for applications like character recognition, image processing, pattern classification etc. One of the drawbacks of the traditional backpropagation method is the need to determine the number of units in the hidden layer prior to training. To overcome this difficulty, many algorithms that construct a network dynamically have been proposed [5]. The most well known constructive algorithms are dynamic node creation (DNC) [1], feedforward neural network construction (FNNC) algorithm [8] and the cascade correlation (CC) algorithm. The DNC algorithm constructs single hidden layer ANNs with a sufficient number of nodes in the hidden layer, though such networks suffers difficulty in learning some complex problems. In contrast, the CC algorithm constructs multiple hidden layer ANNs with one node in each layer and is suitable for some complex problems. However, the CC algorithm has many practical problems, such as difficult implementation in VLSI and long propagation delay [4]. This paper describes, a new algorithm, the modified feedforward neural network constructive algorithm (MFNNCA) for medical diagnosis. It begins network design in a constructive fashion by adding nodes one after another based on the performance of the network on training data.

## 2. NETWORK TOPOLOGY

The size of a feedforward network depends on the number of nodes in the input layer, hidden layer and output layer. The number of nodes in the input layer is defined by the input elements in the input vector; the corresponding output vector defines the number of output nodes in the output layer.

### 2.1 Automatic Determination of Hidden Units with Constructive Approach

Constructive algorithms offer an attractive framework for the incremental construction of near-minimal neural-network architectures. These algorithms start with a small network (usually a single neuron) and dynamically grow the network by adding and training neurons as needed until a satisfactory solution is found. The constructive algorithm proposed in the next section starts with one unit in the hidden layer. Additional units are added to the hidden layer one at a time to improve the accuracy of the network on the training data.

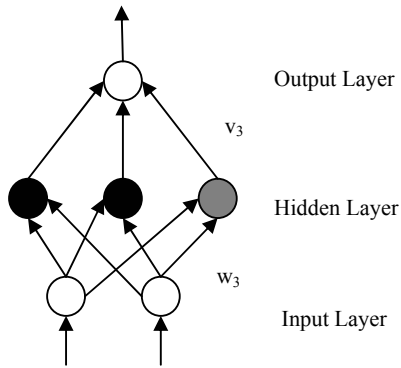

**Fig. 1** Feedforward NN with 3 hidden units.

Figure 1 shows the construction technique of a feedforward NN having 2 initial hidden units, when a third hidden unit is added, the optimal weights obtained from the original network with two hidden units are used as initial weights for retraining the new network [9]. The initial value for $w_3$ and $v_3$ are chosen randomly.

## 3. MODIFIED FEEDFORWARD NEURAL NETWORK CONSTRUCTIVE ALGORITHM (MFNNCA).

The constructive algorithm proposed in this section is based on the feedforward neural network construction algorithm (FNNCA) proposed by Rudy Setiono and Huan Liu [8]. We termed this algorithm as modified FNNCA (MFNNCA). While our algorithm (MFNNCA) is based on FNNCA, it has a significance difference with FNNCA. In FNNCA they have defined the stopping condition of the training by classifying all the input patterns. It means that while the efficiency is 100%, the training will stop. But in most cases with the benchmarking classification problems 100% efficiency may not be achieved. This is why we proposed a new algorithm for pattern classification that defines the stopping condition by the acceptance of efficiency level. Another consideration we have made that the acceptable efficiency on the test sets may not be achieved even though the mean square error on training set is minimum. These considerations encouraged us to propose an algorithm that will combine the learning rule of backpropagation algorithm to update weights of the network and the constructive algorithm to construct the network dynamically and also consider the efficiency factor as a determinant of the training process.

### 3.1 Steps of the Modified Feedforward NN Constructive (MFNNC) Algorithm

To build and train a network the following steps are followed:
1. Create an initial neural network with number of hidden unit $h = 1$. Set all the initial weights of the network randomly within a certain range.
2. Train the network on training set by using a training algorithm for a certain number of epochs that minimizes the error function.
3. If the error function $\xi_{av}$ on validation set is acceptable and, at this position, the network classifies desired number of patterns on test set that leads the efficiency $E$ to be acceptable then *stop*.
4. Add one hidden unit to hidden layer. Randomly initialize the weights of the arcs connecting this new hidden unit with input nodes and output unit(s). Set $h = h + 1$ and go to step 2.

### 3.2 Flow-Chart of the MFNNC Algorithm

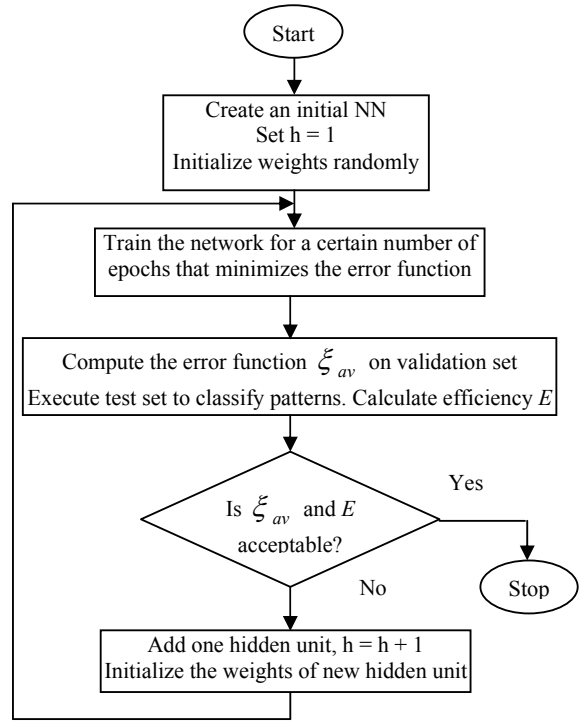

**Fig. 2** Flow chart of the MFNNCA

The error function is usually defined as the mean-squared-errors

$$\xi(n) = \frac{1}{2} \sum_{k \in C} e_k^2(n) \text{------------------(1)}$$

$$\xi_{av} = \frac{1}{N} \sum_{n=1}^{N} \xi(n) \quad \text{----------------(2)}$$

$$e_k = d_k(n) - y_k(n) \quad \text{------------------(3)}$$

Where, k denotes kth output unit, n denotes the nth iteration, C is the number of output units, N is the total number of patterns, $d_k$ denotes the desired output from k, $y_k$ denotes the actual output of neuron k, $e_k$ denotes the error term for kth output unit.

## 4. EXPERIMENTAL RESULTS

This section evaluates the performance of our proposed algorithm (MFNNCA) on three benchmark classification problems i.e, breast cancer (cancer and cancer1), heart disease and diabetes identification. These problems have been subject of many studies in machine learning and neural networks [6]. The data sets representing these problems were obtained from the UCI machine learning benchmark repository and were real world data. Table 1 shows the characteristics of the data sets. The detailed descriptions of the data sets are available at ics.uci.edu. Table 2 shows the experimental results on aforesaid data sets. Figure 3 shows the mean square error curve for cancer data set.

**Table 1:** Characteristics of data sets

| Data set | Input Attributes | Output Units | Output Classes | Training Examples | Validation Examples | Test Examples | Total Examples |
|---|---|---|---|---|---|---|---|
| Cancer | 9 | 1 | 2 | 350 | 175 | 174 | 699 |
| Cancer1 | 9 | 2 | 2 | 350 | 175 | 174 | 699 |
| Heart | 13 | 1 | 2 | 152 | 76 | 75 | 303 |
| Diabetes | 8 | 2 | 2 | 384 | 192 | 192 | 768 |

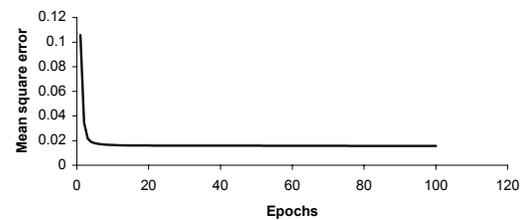

**Fig. 3** Training time error of a network (9-1-1) for Cancer data set

**Table 2:** Experimental results shows the best fitted network by the bolded rows for each data set

| | NO OF HU | EPOCH | Training set | | | Valid set | | | Test set | | OVERALL EFFICIENCY |
|---|---|---|---|---|---|---|---|---|---|---|---|
| | | | CLASSIFIED | EFFICIENCY | MS ERROR | CLASSIFIED | EFFICIENCY | MS ERROR | CLASSIFIED | EFFICIENCY | |
| **Cancer** | 1 | 100 | 338 | 96.57 | 0.0156 | 169 | 96.00% | 0.0171 | 172 | 98.00% | 97.13877 |
| | **2** | **450** | **339** | **96.86** | **0.0138** | **167** | **95.00%** | **0.0229** | **173** | **99.00%** | **97.13877** |
| | 3 | 550 | 339 | 96.86 | 0.0136 | 167 | 95.00% | 0.0229 | 172 | 98.00% | 96.99571 |
| | 4 | 1350 | 346 | 98.86 | 0.008 | 169 | 96.00% | 0.0171 | 168 | 96.00% | 97.71102 |
| **Cancer1** | 1 | 300 | 256 | 73.14 % | 0.0273 | 167 | 95.00 % | 0.0457 | 170 | 97.00 % | 84.83 % |
| | **2** | **600** | **263** | **75.14 %** | **0.0255** | **170** | **97.00 %** | **0.0286** | **171** | **98.00 %** | **86.41 %** |
| | 3 | 700 | 264 | 75.43 % | 0.0253 | 170 | 97.00 % | 0.0286 | 171 | 98.00 % | 86.55 % |
| | 4 | 800 | 264 | 75.43 % | 0.0252 | 170 | 97.00 % | 0.0286 | 171 | 98.00 % | 86.55 % |
| **Heart** | 1 | 500 | 143 | 93.46 | 0.0395 | 64 | 85.00% | 0.0855 | 60 | 80.00% | 88.11881 |
| | 2 | 600 | 144 | 94.12 | 0.0395 | 63 | 84.00% | 0.0855 | 60 | 80.00% | 88.11881 |
| | 3 | 700 | 139 | 90.85 | 0.0428 | 63 | 84.00% | 0.0987 | 60 | 80.00% | 86.46865 |
| | **4** | **800** | **142** | **92.81** | **0.0428** | **62** | **82.00%** | **0.1118** | **61** | **81.00%** | **87.45875** |
| **Diabetes** | 1 | 200 | 300 | 78.12 | 0.2083 | 141 | 73.00% | 0.2656 | 132 | 68.00% | 74.60938 |
| | 2 | 380 | 312 | 81.25 | 0.1875 | 147 | 76.00% | 0.2344 | 135 | 70.00% | 77.34375 |
| | 3 | 480 | 311 | 80.99 | 0.1875 | 147 | 76.00% | 0.2344 | 135 | 70.00% | 77.21354 |
| | **4** | **1580** | **324** | **84.38** | **0.151** | **148** | **77.00%** | **0.2292** | **143** | **74.00%** | **80.07813** |
| | 5 | 1900 | 327 | 85.16 | 0.151 | 147 | 76.00% | 0.2396 | 130 | 67.00% | 74.60938 |

Figure 4 shows the hidden node addition for cancer data set.

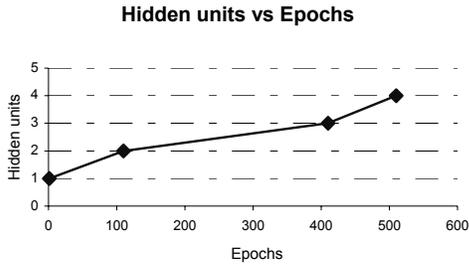

**Fig. 4** Hidden units addition for Cancer data set

## 5. DISCUSSION AND COMPARISON

Medical diagnosis by neural network is the black-box approach: A network is chosen and trained with examples of all classes. After successful training, the system is able to diagnose the unknown cases and to make predictions. For building neural network we have used constructive fashion. Table 3 shows a comparative study of the performance of our algorithm with the works of other researchers.

**Table 3:** Comparative study of the performance

| Data Set | Researchers | % Efficiency | Proposed Algorithm | |
| --- | --- | --- | --- | --- |
| | | | Eff. on test data | Overall Efficiency |
| Cancer | W. H. Wolberg et al. [11] | 93.5 | 99 | 97.14 |
| | R. Setiono [10] | 96 | | |
| Heart | W. David Aha and Dennis Kibler [2] | 77 | 81 | 86.46 |
| | R. Setiono [10] | 83.15 | | |
| Diabetes | J. W., Everhart et al. [3] | 76 | 74 | 80.08 |
| | R. Setiono [10] | 76.32 | | |

## 6. FUTURE WORK

The hidden layer of a neural network plays an important role for detecting the relevant features. Due to the existence of irrelevant and redundant attributes, by selecting only the relevant attributes, higher predictive accuracy can be achieved. For a particular input, any (or few) feature(s) may not be effective to the hidden layer or feature space. By extracting this (these) features we can minimize the training time. In near future, we will try to extend the algorithm for improving backpropagation using feature selection [9].

## 7. CONCLUSION

We have proposed an efficient algorithm (MFNNCA) for medical diagnosis. The novelty of the MFNNCA is that it can determine the number of nodes in a single hidden layer automatically. The training of an artificial neural network by using the traditional backpropagation algorithm may be costly, as the number of hidden units has to be determined prior to training. Introducing constructive algorithms for feedforward networks eliminates the predetermination. As we were focusing at the optimal size of a network that performs the pattern classification with acceptable efficiency, we have ignored all the factors that improve the performance of backpropagation algorithm. The experimental results for the cancer, heart disease and diabetes problems show that the effectiveness of the MFNNCA.